\PassOptionsToPackage{warn}{babel}
\PassOptionsToPackage{numbers,sort&compress}{natbib}
\documentclass{article}
\usepackage{hyphenat}

\usepackage{PRIMEarxiv}

\usepackage[utf8]{inputenc}
\usepackage[T1]{fontenc}
\usepackage{natbib}
\usepackage{hyperref}
\usepackage{url}
\usepackage{booktabs}
\usepackage{amsmath,amsfonts}
\usepackage{nicefrac}
\usepackage{microtype}
\usepackage{fancyhdr}
\usepackage{graphicx}
\usepackage{multirow}
\usepackage{hyphenat}

\title{BUSTR: Descriptor-Aware Vision-Language Learning for Breast Ultrasound Report Generation
}

\author{
  Rawa Mohammed, Mina Attin, Laxmi Gewali, Bryar Shareef \\
  University of Nevada, Las Vegas
}

\begin{document}
\maketitle

\begin{abstract}
Breast ultrasound (BUS) reporting relies on clinically meaningful lesion descriptors, including BI-RADS category, lesion shape, margin, echogenicity, posterior features, pathology, and histology. However, many public BUS datasets provide structured annotations and lesion masks without paired radiologist-written reports, limiting the development of vision--language models for BUS report generation. We propose BUSTR, a descriptor-aware vision--language framework that uses structured lesion information to enable report generation under limited report supervision. BUSTR first constructs descriptor-derived reports from available annotations and radiomics features extracted from lesion masks. It then trains a multi-head Swin Transformer encoder with multitask supervision to learn descriptor-aware visual representations across datasets with partially overlapping annotation sets. The projected visual tokens condition a frozen LLaMA-based language model, and training is guided by a dual-level objective combining token-level cross-entropy with representation-level cosine alignment. At inference, BUSTR generates reports from BUS images without access to structured descriptors, lesion masks, or radiomics features. We evaluate BUSTR on the public BrEaST and BUS-BRA datasets using natural language generation and clinical efficacy metrics. BUSTR improves report similarity and descriptor recovery compared with representative report-generation baselines, with notable gains for lesion shape, margin, posterior features, and pathology, as well as improved BI-RADS sensitivity and F1-score on BrEaST. These results suggest that structured BUS descriptors, lesion masks, and radiomics features can provide useful supervision for descriptor-aware BUS report generation when paired radiologist-written reports are unavailable.
\end{abstract}


\section{Introduction}

Breast ultrasound (BUS) is an important imaging modality for breast lesion assessment, particularly in patients with dense breast tissue, where mammography may have reduced sensitivity \cite{kolb2002comparison}. BUS is widely used because it is accessible, relatively low-cost, and does not involve ionizing radiation. However, interpretation of BUS images remains challenging because of operator dependency, speckle noise, low contrast, artifacts, and variability in lesion appearance \cite{busis,shareef2022estan}. At the same time, the growing demand for breast imaging has increased radiology workload and contributed to delays in interpretation and reporting \cite{markit2017complexities}. These challenges motivate artificial intelligence (AI) systems that can extract clinically meaningful information from BUS images and organize it into structured report text.

Most existing BUS computer-aided diagnosis (CAD) systems focus on lesion segmentation \cite{sun2025challenge,shareef2022estan} or classification \cite{mahesh2025enhancing}. Curated benchmarks and tumor-aware architectures have improved BUS image analysis and supported the development of downstream diagnostic models \cite{busis}. Recent multitask learning approaches have further improved representation quality by jointly modeling tumor classification with auxiliary clinical or imaging tasks \cite{BCS-NET,ChenAlign,shareef2023hybrid}. Although these methods are valuable for decision support, their outputs are often limited to segmentation masks, class probabilities, BI-RADS categories, or isolated quantitative predictions. Such outputs can support triage and risk stratification, but they do not directly organize image-derived findings into clinically meaningful report text.

Structured lesion descriptors play a central role in BUS assessment and reporting. BI-RADS provides a standardized vocabulary for describing breast imaging findings and supports structured reporting and data mining \cite{zhang2023bi,margolies2016_bigdata}. Public BUS datasets often include annotations such as BI-RADS category, lesion shape, margin, echogenicity, posterior features, pathology labels, histology labels, lesion masks, or related metadata. These annotations can serve not only as targets for classification, but also as intermediate supervision for learning clinically meaningful visual representations. In addition, lesion masks enable the extraction of radiomics features that provide quantitative information about lesion size, intensity, and morphology. Together, structured descriptors, masks, and radiomics features provide a useful source of supervision for BUS report generation.

Despite this potential, many public BUS datasets do not include paired radiologist-written narrative reports. This limits the direct use of conventional image-to-report generation methods, which are commonly trained on paired image--report corpora such as chest radiography datasets. Automated radiology report generation (RRG) has advanced from template-based and retrieval-based methods \cite{tommasi2008svm,gobeill2009query} to deep learning approaches that combine visual encoders with recurrent or Transformer-based language models \cite{sirshar2022attention,wang2025survey,wang2023_metransformer}. More recently, large language models (LLMs) have been introduced into radiology report generation because of their strong language-modeling ability and capacity to produce fluent and context-aware text \cite{sun2023evaluating,wang2023r2gengpt}. However, LLM-based medical report generation can produce unsupported or hallucinated findings when visual grounding is weak or supervision is limited \cite{pal2024gemini,ramesh2022improving}. Recent methods have attempted to improve factual consistency through visual-concept alignment \cite{gu2025radalign}, reinforcement learning, or text augmentation \cite{parres2024_rrg_rl}, but these approaches generally still depend on paired image--report datasets.

Several studies have investigated AI-assisted BUS diagnosis and report generation \cite{ge2023ai,li2024ultrasound,lo2024automated,qin2023computer,azhar2024_bus_reports,huh2025wholistic}. Qin et al.\ \cite{qin2023computer} proposed a deep learning-based system for BUS report generation and classification, but the primary emphasis was lesion categorization. Lo and Chen \cite{lo2024automated} introduced a metadata-driven reporting system that integrates multiple imaging features for shared decision-making, although the generated outputs remain strongly dependent on structured metadata. Li et al.\ \cite{li2024ultrasound} proposed an ultrasound report generation method based on cross-modality feature alignment with unsupervised guidance, but its reliance on private data limits reproducibility and direct comparison with public BUS datasets. Azhar et al.\ \cite{azhar2024_bus_reports} synthesized structured multimodal BUS reports by combining radiologist annotations with deep learning analysis, while Huh et al.\ \cite{huh2025wholistic} integrated multiple BUS analysis tools within a LangChain-based LLM framework. Ge et al.\ \cite{ge2023ai} also incorporated BI-RADS descriptors into structured BUS reporting, but the generated outputs were primarily template-based. These studies demonstrate growing interest in BUS reporting support, but many rely on private datasets, radiologist-written annotations, metadata-driven pipelines, or template-style generation.

More broadly, recent medical report generation methods have introduced disease-aware \cite{park2025dart}, abnormal-region-aware \cite{gao2025abnormal}, organ-aware \cite{li2024organ}, and anatomy-aware \cite{chen2024anatomy} representations to improve image--text alignment. These studies show that clinically meaningful intermediate representations can improve report generation. However, they are mainly designed for settings in which paired reports are available. BUS presents a different and practical setting: structured lesion descriptors and masks are often available, whereas paired narrative reports are not. This motivates a descriptor-aware learning strategy that uses available structured annotations and radiomics features to support visual representation learning and report generation.

In this work, we propose BUSTR, a descriptor-aware vision--language framework for BUS report generation. Instead of treating the absence of paired radiologist-written reports as a complete barrier, BUSTR uses structured BUS descriptors and radiomics features to construct descriptor-derived reference reports, hereafter referred to as descriptor-derived reports. These reports are constrained to restate available descriptor values and image-derived quantitative features in a clinically organized format. They are not intended to replace expert-authored clinical reports; rather, they provide fact-based textual supervision for learning to connect BUS image features with structured report content.

BUSTR first constructs descriptor-derived reports from available structured annotations and radiomics features extracted from lesion masks. It then trains a descriptor-aware multi-head Swin Transformer encoder using multitask supervision across BUS datasets with partially overlapping annotation sets. Finally, projected visual tokens condition a frozen LLaMA-based language model, and training is guided by a dual-level objective that combines token-level cross-entropy with representation-level cosine alignment. At inference, BUSTR generates reports from BUS images without access to structured descriptors, lesion masks, radiomics features, or descriptor-derived target reports.

The main contributions of this work are as follows:
\begin{enumerate}
    \item A descriptor-aware vision--language framework for breast ultrasound report generation using structured lesion annotations, lesion masks, and radiomics features.

    \item A descriptor-derived report construction strategy that enables report-level supervision when paired radiologist-written reports are unavailable.

    \item A multitask Swin Transformer encoder that learns descriptor-aware BUS visual representations across datasets with partially overlapping annotation sets.

    \item A dual-level vision--language training objective that combines token-level cross-entropy with representation-level alignment to improve report similarity and descriptor recovery.
\end{enumerate}

\begin{figure}[h]
\centering
\includegraphics[width=0.8\textwidth]{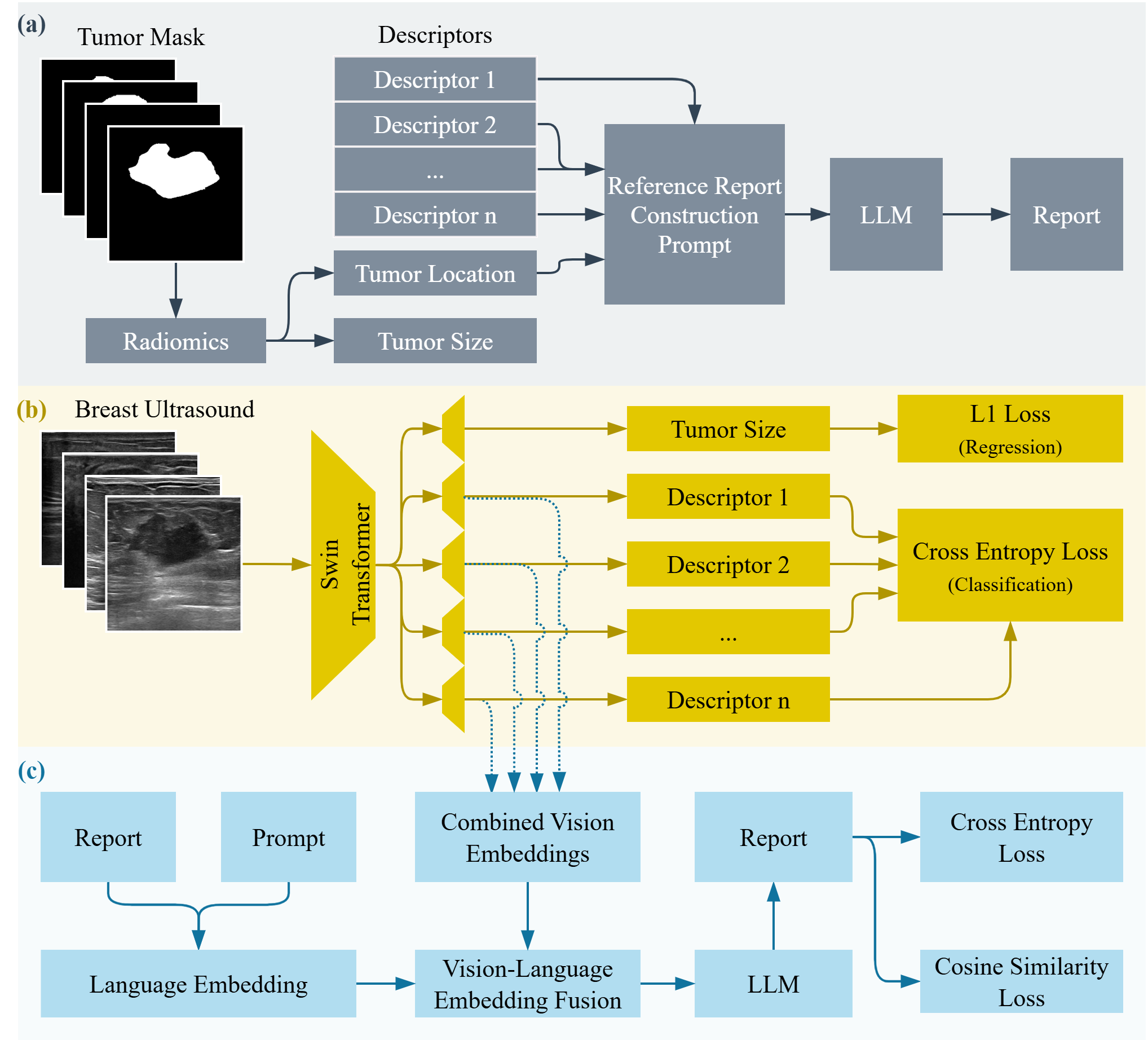}
\caption{Overall BUSTR architecture.
(a) Descriptor-derived report construction from BUS descriptors and
radiomics features.
(b) Multitask training of a descriptor-aware Swin vision encoder.
(c) Report generation, where projected visual tokens and a fixed prompt
condition a frozen LLM to produce a BUS report. The illustrated descriptor
heads correspond to the BrEaST configuration; the active output heads are
adapted to the annotations available in BUS-BRA.}
\label{fig:BARG}
\end{figure}

\section{Proposed Method}

\subsection{Overview}

The proposed framework, BUSTR, is designed to connect breast ultrasound (BUS) image features with structured lesion descriptors and clinically organized report text. As shown in Fig.~\ref{fig:BARG}, BUSTR consists of three main components: (1) descriptor-derived report construction, where available structured annotations and radiomics features are converted into constrained report-level supervision; (2) descriptor-aware visual representation learning, where a multitask Swin Transformer encoder is trained to predict multiple lesion descriptors and auxiliary targets from BUS images; and (3) vision--language report generation, where projected visual tokens and textual prompts are used to generate BUS reports grounded in image-derived lesion information.

The same framework is applied to both datasets, but the active prediction heads and supervision depend on the annotations available in each dataset. The architecture illustrated in Fig.~\ref{fig:BARG} corresponds to the BrEaST configuration, while the output heads are adapted to the annotation set available in BUS-BRA. This design allows BUSTR to learn from public BUS datasets with partially overlapping annotations while maintaining a unified descriptor-aware report-generation pipeline.

\subsection{Descriptor-Derived Report Construction}

Public BUS datasets often provide structured lesion annotations and lesion masks, but not paired radiologist-written narrative reports. To create report-level supervision under this constraint, we construct descriptor-derived reference reports, hereafter referred to as descriptor-derived reports, from structured BUS descriptors and radiomics features. A LLaMA-based model~\cite{touvron2023llama} is used to convert structured, fact-based information into concise report-style text.

The available descriptors may include BI-RADS category, lesion shape, margin, echogenicity, posterior features, histology, pathology, and radiomics features extracted from tumor masks, such as lesion size and intensity statistics. Because different datasets provide different annotation sets, only the descriptors available for a given image are used.

The descriptor-derived reports are constrained to restate the available structured information and radiomics-derived attributes. They are not intended to introduce new findings or replace expert-authored clinical reports. This design provides a consistent supervisory signal for linking BUS image features with clinically organized text while reducing unsupported or hallucinated content.

To improve factual consistency during report construction, we used a constrained prompt template that was iteratively refined through qualitative inspection of generated texts. Early prompt variants sometimes produced unsupported content, including speculative disease statements, repeated descriptors, treatment suggestions, malignancy percentages, demographic details not present in the input, or overly informal phrasing. We therefore added explicit constraints requiring the language model to use only the provided descriptors, avoid repetition, maintain concise formal language, and exclude unsupported clinical or demographic information.

For a given BUS image, we collect all available descriptor--value pairs into
\[
y_{\text{descriptors}} = \{(d, y_d)\},
\]
where \(d\) denotes a descriptor type, such as BI-RADS category, shape, margin, histology, or pathology, and \(y_d\) is the corresponding descriptor value. Radiomics features are extracted from the tumor mask as
\[
\mathbf{R} = f_{\text{rad}}(M),
\]
where \(M\) is the tumor mask and \(f_{\text{rad}}(\cdot)\) is a fixed radiomics feature extractor.

A formatting function \(\mathcal{F}_{\text{format}}\) converts the available descriptors and radiomics features into an instruction-style prompt:
\begin{equation}
    P = \mathcal{F}_{\text{format}}(y_{\text{descriptors}}, \mathbf{R}),
\end{equation}
where \(P\) contains only the structured information available for that image. The frozen LLM then generates the descriptor-derived report:
\begin{equation}
    R_{\text{ref}} = \mathcal{F}_{\text{LLM}}(P).
\end{equation}
This process is applied to all images, with the descriptor set varying according to the annotations available in each dataset.

\subsection{Descriptor-Aware Vision Encoder}
\label{sec:vision_encoder}

The visual representation learning component uses a pre-trained Swin Transformer~\cite{liu2021swin} as the BUS image encoder. The encoder is trained in a multitask setting to predict clinically meaningful lesion descriptors and auxiliary targets. Let \(I \in \mathbb{R}^{H \times W}\) be an input BUS image, and let
\[
\mathbf{E}_v = f_{\text{enc}}(I) \in \mathbb{R}^{N_p \times D}
\]
denote the encoder output, where \(N_p\) is the number of image patches and \(D\) is the embedding dimension.

The Swin encoder is followed by multiple task-specific prediction branches. The active branches depend on the annotations available in each dataset. For BrEaST, the prediction targets include BI-RADS category, lesion shape, margin, posterior features, echogenicity, and tumor size. For BUS-BRA, the prediction targets include BI-RADS category, pathology, histology, and tumor size. The corresponding task sets are
\begin{equation}
\begin{aligned}
\mathcal{T}_{\text{BrEaST}}
&=
\{
\text{BI-RADS},
\text{shape},
\text{margin},
\text{posterior},
\text{echogenicity},
\text{size}
\},\\
\mathcal{T}_{\text{BUS-BRA}}
&=
\{
\text{BI-RADS},
\text{pathology},
\text{histology},
\text{size}
\}.
\end{aligned}
\end{equation}

For each active branch, adaptive average pooling is applied over the patch dimension to obtain a fixed-length hidden representation, such as \(\mathbf{h}_{\text{shape}}\), \(\mathbf{h}_{\text{margin}}\), \(\mathbf{h}_{\text{posterior}}\), \(\mathbf{h}_{\text{echo}}\), \(\mathbf{h}_{\text{pathology}}\), \(\mathbf{h}_{\text{histology}}\), or \(\mathbf{h}_{\text{size}}\). These representations are then provided to the corresponding prediction heads.

For BrEaST, BI-RADS prediction is implemented as a descriptor-aware classifier that uses the joint representation of the four available lesion descriptors:
\[
\hat{y}_{\text{BI-RADS}} =
h_{\text{BI-RADS}}
\big(
\text{concat}(
\mathbf{h}_{\text{shape}},
\mathbf{h}_{\text{margin}},
\mathbf{h}_{\text{posterior}},
\mathbf{h}_{\text{echo}}
)
\big).
\]
This design is motivated by the role of lesion shape, margin, posterior features, and echogenicity in BI-RADS assessment. For BUS-BRA, separate task-specific classification heads are used for BI-RADS category, pathology, and histology because the detailed lesion descriptors available in BrEaST are not provided in BUS-BRA.

The complete set of possible task-specific predictions is expressed as
\begin{equation}
\begin{aligned}
\hat{y}_{\text{shape}}     &= h_{\text{shape}}(\mathbf{h}_{\text{shape}}), \\
\hat{y}_{\text{margin}}    &= h_{\text{margin}}(\mathbf{h}_{\text{margin}}), \\
\hat{y}_{\text{posterior}} &= h_{\text{posterior}}(\mathbf{h}_{\text{posterior}}), \\
\hat{y}_{\text{echo}}      &= h_{\text{echo}}(\mathbf{h}_{\text{echo}}), \\
\hat{y}_{\text{pathology}} &= h_{\text{pathology}}(\mathbf{h}_{\text{pathology}}), \\
\hat{y}_{\text{histology}} &= h_{\text{histology}}(\mathbf{h}_{\text{histology}}), \\
\hat{y}_{\text{size}}      &= h_{\text{size}}(\mathbf{h}_{\text{size}}),
\end{aligned}
\end{equation}
where only the heads corresponding to the annotations available in the current dataset are active. The tumor-size prediction \(\hat{y}_{\text{size}}\) is a normalized scalar. Tumor size is scaled by the maximum value during training, consistent with the implementation.

During inference, the tumor-size head provides a quantitative size estimate that can be inserted into the generated report. Empirically, adding tumor size as an additional visual token for the language model did not improve performance; therefore, tumor size is not included in the visual token sequence.

\subsubsection*{Multitask Descriptor Loss}

For each supervised target, a separate loss term is defined. Categorical descriptors are optimized using cross-entropy loss, while tumor size is optimized using an L1 loss on the normalized size:
\begin{equation}
\begin{aligned}
\mathcal{L}_{\text{BI-RADS}}
&= \text{CE}(\hat{y}_{\text{BI-RADS}}, y_{\text{BI-RADS}}), \\
\mathcal{L}_{\text{shape}}
&= \text{CE}(\hat{y}_{\text{shape}}, y_{\text{shape}}), \\
\mathcal{L}_{\text{margin}}
&= \text{CE}(\hat{y}_{\text{margin}}, y_{\text{margin}}), \\
\mathcal{L}_{\text{posterior}}
&= \text{CE}(\hat{y}_{\text{posterior}}, y_{\text{posterior}}), \\
\mathcal{L}_{\text{echo}}
&= \text{CE}(\hat{y}_{\text{echo}}, y_{\text{echo}}), \\
\mathcal{L}_{\text{pathology}}
&= \text{CE}(\hat{y}_{\text{pathology}}, y_{\text{pathology}}), \\
\mathcal{L}_{\text{histology}}
&= \text{CE}(\hat{y}_{\text{histology}}, y_{\text{histology}}), \\
\mathcal{L}_{\text{size}}
&= \text{L1}(\hat{y}_{\text{size}}, y_{\text{size}}).
\end{aligned}
\end{equation}
Only the loss terms corresponding to annotations available in the current dataset are included in the multitask objective.

Some descriptors contain multiple related labels. In particular, margin includes a main class, circumscribed versus non-circumscribed, and four non-circumscribed subtypes: indistinct, angular, spiculated, and microlobulated. Let \(\mathcal{L}_{\text{Margin}}\) denote the loss for the main margin class, and let \(\mathcal{L}_{\text{margin},j}\) denote the loss for subtype \(j\), where \(j \in \{1,\dots,4\}\). The combined margin loss is defined as
\begin{equation}
\mathcal{L}_{\text{Combined Margin}}
= 0.5\,\mathcal{L}_{\text{Margin}}
+ \frac{0.5}{4} \sum_{j=1}^{4} \mathcal{L}_{\text{margin},j}.
\end{equation}
Thus, the main margin class and the mean of the four subtype losses contribute equally.

Let \(\mathcal{L}_t\) denote the loss for task \(t\) in the active supervised task set for a given training configuration, and let \(w_t\) denote the corresponding task weight. The overall multitask vision loss is
\begin{equation}
\mathcal{L}_{\text{Vision}}^{(\text{cfg})}
=
\sum_{t \in \mathcal{T}^{(\text{cfg})}} w_t \mathcal{L}_t,
\end{equation}
where \(\mathcal{T}^{(\text{cfg})}\) denotes the set of active tasks for that configuration. For BrEaST, BI-RADS and tumor size are each assigned 20\% of the total multitask vision loss, while shape, margin, posterior features, and echogenicity are each assigned 15\%. For BUS-BRA, multiple task-weight configurations are evaluated in the sensitivity analysis, and the configuration with the best overall report generation performance is selected. This formulation allows the model to use different descriptor sets across datasets while preserving a consistent descriptor-aware training strategy.

\subsection{Vision--Language Report Generation}

The report generation component uses the descriptor-aware vision encoder to condition a frozen LLaMA-based language model. The goal is to generate a BUS report that reflects image-derived lesion information and remains consistent with the structured descriptor supervision.

During training, the word embedding sequence is produced by the LLM embedding model \(\mathbf{E}_{\text{LLM}}\) from the descriptor-derived report \(T\) and the instruction-style prompt \(P\):
\[
\mathbf{E}_{\text{word}} = \mathbf{E}_{\text{LLM}}(T, P).
\]
The vision encoder produces visual tokens \(\mathbf{E}_{v}\) from the BUS image, and these tokens are projected into the LLaMA embedding space. Let \(\tilde{\mathbf{E}}_{v}\) denote the projected visual tokens. The projected visual tokens are concatenated with the word embeddings and passed to the frozen LLM:
\begin{equation}
    R_{\text{final}} =
    \mathcal{F}_{\text{LLM}}
    \big(
    [\tilde{\mathbf{E}}_{v}; \mathbf{E}_{\text{word}}]
    \big).
\end{equation}

Clinical terms in BUS reports, such as BI-RADS descriptors, lesion size, margin, posterior features, pathology, and histology, are more important than general words. Many of these terms may be split into multiple sub-tokens by standard tokenizers, increasing the chance of incomplete or unsupported term generation. To reduce this issue, relevant clinical terms are added to the tokenizer as independent tokens before training, and the LLM embedding matrix is resized accordingly.

The language model is kept frozen because of the limited dataset size, while the vision encoder is fine-tuned from the multitask descriptor-aware weights. The proposed design does not rely on a specific narrative style of the selected LLM; instead, the LLM is used as a general text generator conditioned on image-derived visual tokens and descriptor-based textual prompts. Although LLaMA is used in this study, the framework can in principle be adapted to other decoder-style language models with compatible token embeddings. Quantitative performance may vary across language models because tokenization behavior and embedding spaces differ.

\subsubsection*{Vision--Language Training Objective}

Training is optimized using two complementary losses: a token-level cross-entropy loss and a representation-level alignment loss. The cross-entropy loss encourages the generated report to match the descriptor-derived report at the token level:
\begin{equation}
\mathcal{L}_{\text{CE}} = -\sum_{i=1}^{L}\log p(x_i),
\end{equation}
where \(p(x_i)\) is the predicted probability of token \(x_i\) at position \(i\), and \(L\) is the report length.

Let \(\mathbf{H} \in \mathbb{R}^{L_z \times D}\) denote the last hidden state of the LLM, and let \(\mathbf{Z} \in \mathbb{R}^{L_z \times D}\) denote the corresponding input embeddings, including visual tokens, prompt tokens, and report tokens. The alignment loss is defined as
\begin{equation}
\mathcal{L}_{\text{Align}} =
1 - \frac{1}{L_z}
\sum_{i=1}^{L_z}
\frac{
\mathbf{H}_i \cdot \mathbf{Z}_i
}{
\|\mathbf{H}_i\| \, \|\mathbf{Z}_i\|
}.
\end{equation}
This loss encourages the final LLM representations to remain aligned with the image-conditioned input embeddings.

The final training objective for the report generation stage is
\begin{equation}
\mathcal{L} =
\lambda_{\text{CE}}\mathcal{L}_{\text{CE}}
+
\lambda_{\text{Align}}\mathcal{L}_{\text{Align}},
\end{equation}
where \(\lambda_{\text{CE}} = \lambda_{\text{Align}} = 0.5\) in our experiments. The combined objective encourages token-level report accuracy while maintaining consistency between the visual--textual inputs and the generated report representations.

\begin{table}[t]
\centering
\caption{BrEaST case distribution by descriptor category.}
\label{tab_distribution_breast}
\small
\renewcommand{\arraystretch}{1.15}
\setlength{\tabcolsep}{0.9em}
\begin{tabular*}{\textwidth}{@{\extracolsep{\fill}}llcllc}
\toprule
Descriptor & Category & Cases & Descriptor & Category & Cases \\
\midrule
\multirow{6}{*}{BI-RADS}
    & 2  & 30 & \multirow{6}{*}{Margin}
    & Circumscribed & 115 \\
    & 3  & 37 & 
    & Non-circumscribed & 137 \\
    & 4A & 44 & 
    & \quad Angular & 42 \\
    & 4B & 46 & 
    & \quad Indistinct & 115 \\
    & 4C & 49 & 
    & \quad Microlobulated & 36 \\
    & 5  & 46 & 
    & \quad Spiculated & 33 \\
\midrule
\multirow{3}{*}{Shape}
    & Oval & 97 & \multirow{6}{*}{Echogenicity}
    & Anechoic & 15 \\
    & Round & 15 & 
    & Hypoechoic & 148 \\
    & Irregular & 140 & 
    & Hyperechoic & 9 \\
\cmidrule(lr){1-3}
\multirow{4}{*}{Posterior features}
    & None & 159 & 
    & Isoechoic & 12 \\
    & Enhancement & 36 & 
    & Heterogeneous & 57 \\
    & Shadowing & 50 & 
    & Complex cystic/solid & 11 \\
    & Combined features & 7 & 
    &  &  \\
\bottomrule
\end{tabular*}
\end{table}

\section{Experiments}

\subsection{Datasets and preprocessing}
\label{sec:datasets}

We evaluate BUSTR on two publicly available breast ultrasound datasets:
BrEaST~\cite{pawlowska2024curated} and BUS-BRA~\cite{gomez2024bus}. Both
datasets provide BUS images and structured annotations that are used to
construct descriptor-derived reports and train the descriptor-aware vision
encoder.

BrEaST contains 256 BUS images with detailed BI-RADS descriptors, including
BI-RADS category, lesion shape, margin with non-circumscribed subtypes,
echogenicity, and posterior features. The dataset also includes curated lesion
masks that enable radiomics feature extraction. However, the number of cases is
relatively small, and several descriptor categories are highly imbalanced. For
report generation, we use five categorical descriptors: BI-RADS category,
shape, margin, echogenicity, and posterior features. Mask-derived tumor size is
additionally used as an auxiliary regression target in the multitask vision
encoder. Other descriptors, including tissue composition, skin thickening, and
calcification, were excluded because of severe class imbalance or a large
number of missing values. Cases with BI-RADS category 1, comprising four
images, were excluded from the analysis. The descriptor distribution is
summarized in Table~\ref{tab_distribution_breast}.

BUS-BRA contains 1{,}875 BUS images and provides BI-RADS category, pathology
(benign vs.\ malignant), and histology annotations. These three categorical
descriptors are used in descriptor-derived report construction and as
classification targets in the multitask vision encoder. Mask-derived tumor
size is additionally included as an auxiliary regression target. The
categorical descriptor distribution for BUS-BRA is presented in
Table~\ref{tab_distribution_busbra}.

For both datasets, images are padded to obtain square inputs and resized to
$224 \times 224$ pixels. We use five-fold cross-validation, and the training
portion of each fold is further divided into 80\% training and 20\%
validation. The same folds are used for all compared methods. No data
augmentation is applied.

\begin{table}[t]
\centering
\caption{BUS-BRA descriptor distribution.}
\label{tab_distribution_busbra}
\small
\renewcommand{\arraystretch}{1.15}
\setlength{\tabcolsep}{0.8em}
\begin{tabular}{llc}
\toprule
Descriptor & Category & Cases \\
\midrule
\multirow{4}{*}{BI-RADS}
    & 2 & 562 \\
    & 3 & 463 \\
    & 4 & 693 \\
    & 5 & 157 \\
\midrule
\multirow{2}{*}{Pathology}
    & Benign & 1268 \\
    & Malignant & 607 \\
\midrule
\multirow{10}{*}{\shortstack[l]{Histology\\Top 10 of 28}}
    & Fibroadenoma & 835 \\
    & Invasive ductal carcinoma & 520 \\
    & Cyst & 142 \\
    & Fibrocystic changes & 106 \\
    & Invasive lobular carcinoma & 42 \\
    & Intraductal papilloma & 41 \\
    & Sclerosing adenosis & 37 \\
    & Hyperplasia & 31 \\
    & Lipoma & 17 \\
    & Phyllodes tumor & 13 \\
\bottomrule
\end{tabular}
\end{table}

\subsection{Training, Inference, and Implementation Details}

\subsubsection{Training stages.}
Training consists of two stages:
\begin{enumerate}
    \item \textbf{Descriptor-aware vision encoder.}
    We first train the Swin Transformer encoder with the task-specific
    prediction heads defined in Sec.~\ref{sec:vision_encoder}. For BrEaST,
    the active targets are BI-RADS category, lesion shape, margin,
    echogenicity, posterior features, and tumor size. For BUS-BRA, the active
    targets are BI-RADS category, pathology, histology, and tumor size. The
    encoder is trained for 100 epochs with a batch size of 8 and a learning
    rate of $1 \times 10^{-4}$ using the multitask vision objective described
    in Sec.~\ref{sec:vision_encoder}.

    \item \textbf{Report generation.}
    In the second stage, the LLaMA language model is kept frozen, while the
    vision encoder is fine-tuned from the multitask weights obtained in
    Stage~1. The report-generation model is trained for 25 epochs on BrEaST
    and 35 epochs on BUS-BRA using a batch size of 8 and a learning rate of
    $1 \times 10^{-4}$. Training uses the combined objective based on
    token-level cross-entropy and representation-level cosine alignment.
\end{enumerate}

The same five-fold partitions are used in both training stages. During
inference, BUSTR receives only a BUS image and a fixed generation prompt.
Structured descriptors, lesion masks, radiomics features, and
descriptor-derived target reports are not provided to the model at inference.

\begin{table}[t]
\centering
\caption{Overall performance using NLG metrics. Bold indicates the best result, and underlining indicates the second-best result.}
\label{tab_nlg}
\small
\renewcommand{\arraystretch}{1.15}
\setlength{\tabcolsep}{0.55em}
\begin{tabular*}{\textwidth}{@{\extracolsep{\fill}}llccccccc}
\toprule
Dataset & Approach & BLEU-1$\uparrow$ & BLEU-2$\uparrow$ & BLEU-3$\uparrow$ & BLEU-4$\uparrow$ & ROUGE-L$\uparrow$ & METEOR$\uparrow$ & CIDEr$\uparrow$ \\
\midrule
\multirow{6}{*}{BrEaST}
& R2GenGPT & 0.586 & 0.465 & 0.383 & 0.324 & 0.442 & 0.290 & 0.159 \\
& R2GenCMN & 0.635 & 0.520 & 0.439 & 0.379 & 0.498 & 0.306 & 0.418 \\
& Li & 0.628 & 0.516 & 0.439 & 0.381 & 0.497 & 0.303 & 0.378 \\
& TSGET & 0.633 & 0.523 & 0.445 & 0.386 & \underline{0.508} & 0.307 & 0.438 \\
& R2Gen & \underline{0.639} & \underline{0.528} & \underline{0.450} & \underline{0.391} & 0.504 & \underline{0.310} & \underline{0.492} \\
& BUSTR & \textbf{0.671} & \textbf{0.558} & \textbf{0.478} & \textbf{0.418} & \textbf{0.554} & \textbf{0.337} & \textbf{0.625} \\
\midrule
\multirow{6}{*}{BUS-BRA}
& R2GenGPT & 0.705 & 0.595 & 0.517 & 0.458 & 0.593 & 0.359 & \underline{1.673} \\
& R2GenCMN & \underline{0.722} & 0.617 & 0.541 & 0.483 & 0.611 & 0.370 & 1.414 \\
& Li & 0.721 & \underline{0.619} & \underline{0.543} & \underline{0.484} & \underline{0.617} & 0.369 & 1.302 \\
& TSGET & 0.718 & 0.610 & 0.533 & 0.475 & 0.602 & \underline{0.371} & 1.484 \\
& R2Gen & 0.715 & 0.607 & 0.531 & 0.472 & 0.606 & 0.365 & 1.298 \\
& BUSTR & \textbf{0.735} & \textbf{0.641} & \textbf{0.571} & \textbf{0.516} & \textbf{0.658} & \textbf{0.398} & \textbf{2.062} \\
\bottomrule
\end{tabular*}
\end{table}

\subsubsection{Implementation details.}
All experiments are conducted on a single machine with an Intel Xeon Gold 6326 CPU
(2.90\,GHz) and NVIDIA A100 80\,GB GPUs. Training all five folds requires
approximately 4 hours for BrEaST and 16 hours for BUS-BRA.

\subsection{Evaluation Metrics} We evaluate the generated reports using two groups of metrics: Natural Language Generation (NLG) metrics and Clinical Efficacy (CE) metrics. \subsubsection{NLG metrics.} To measure similarity between the generated reports and the descriptor-derived reports, we use BLEU-1 through BLEU-4~\cite{papineni2002bleu}, ROUGE-L~\cite{lin2004rouge}, METEOR~\cite{banerjee2005meteor}, and CIDEr~\cite{vedantam2015cider}. These metrics assess complementary aspects of lexical overlap and report-level similarity, including n-gram correspondence, longest common subsequences, and consensus-based agreement. \subsubsection{CE metrics.} To assess preservation of clinically relevant content, we compute precision (P), sensitivity (S), and F1-score (F1) for selected lesion descriptors and BI-RADS categories recovered from the generated reports. For BrEaST, CE metrics are reported for BI-RADS, shape, echogenicity, margin, and posterior features. For BUS-BRA, BI-RADS, pathology, and histology are evaluated. These metrics assess whether the generated reports preserve the structured lesion information represented in the descriptor-derived targets. 
\subsection{Overall Performance}
\begin{table}[t]
\centering
\caption{Overall performance using CE metrics for the BrEaST dataset. Bold indicates the best result, and underlining indicates the second-best result.}
\label{tab_ce_breast}
\small
\renewcommand{\arraystretch}{1.15}
\setlength{\tabcolsep}{0.65em}
\begin{tabular*}{\textwidth}{@{\extracolsep{\fill}}llcccccc}
\toprule
Descriptor & Metric & R2GenGPT & R2GenCMN & Li & R2Gen & TSGET & BUSTR \\
\midrule
\multirow{3}{*}{BI-RADS}
& P$\uparrow$  & 0.166 & 0.178 & 0.105 & \textbf{0.329} & 0.218 & \underline{0.288} \\
& S$\uparrow$  & 0.115 & 0.235 & 0.171 & \underline{0.250} & 0.206 & \textbf{0.314} \\
& F1$\uparrow$ & 0.123 & 0.182 & 0.114 & \underline{0.244} & 0.165 & \textbf{0.274} \\
\midrule
\multirow{3}{*}{Shape}
& P$\uparrow$  & 0.490 & 0.559 & 0.602 & \underline{0.689} & 0.655 & \textbf{0.704} \\
& S$\uparrow$  & 0.457 & 0.592 & 0.595 & \underline{0.654} & 0.638 & \textbf{0.711} \\
& F1$\uparrow$ & 0.447 & 0.529 & 0.577 & \underline{0.647} & 0.613 & \textbf{0.686} \\
\midrule
\multirow{3}{*}{Echogenicity}
& P$\uparrow$  & 0.348 & 0.384 & 0.344 & \underline{0.415} & \textbf{0.422} & 0.349 \\
& S$\uparrow$  & \underline{0.560} & \textbf{0.568} & 0.556 & 0.528 & \textbf{0.568} & \underline{0.560} \\
& F1$\uparrow$ & 0.428 & 0.445 & 0.424 & \textbf{0.453} & \underline{0.449} & 0.428 \\
\midrule
\multirow{3}{*}{Margin}
& P$\uparrow$  & 0.631 & 0.612 & 0.645 & \underline{0.757} & 0.726 & \textbf{0.818} \\
& S$\uparrow$  & 0.523 & 0.612 & 0.639 & \underline{0.710} & 0.678 & \textbf{0.794} \\
& F1$\uparrow$ & 0.496 & 0.559 & 0.629 & \underline{0.711} & 0.666 & \textbf{0.795} \\
\midrule
\multirow{3}{*}{Posterior features}
& P$\uparrow$  & 0.401 & 0.451 & 0.401 & \underline{0.556} & 0.427 & \textbf{0.587} \\
& S$\uparrow$  & 0.592 & 0.631 & 0.631 & \underline{0.635} & 0.572 & \textbf{0.659} \\
& F1$\uparrow$ & 0.478 & 0.517 & 0.490 & \underline{0.556} & 0.477 & \textbf{0.566} \\
\bottomrule
\end{tabular*}
\end{table}

\begin{table}[t]
\centering
\caption{Overall performance using CE metrics for the BUS-BRA dataset. Bold indicates the best result, and underlining indicates the second-best result.}
\label{tab_ce_busbra}
\small
\renewcommand{\arraystretch}{1.15}
\setlength{\tabcolsep}{0.65em}
\begin{tabular*}{\textwidth}{@{\extracolsep{\fill}}llcccccc}
\toprule
Descriptor & Metric & R2GenGPT & R2GenCMN & Li & TSGET & R2Gen & BUSTR \\
\midrule
\multirow{3}{*}{BI-RADS}
& P$\uparrow$  & 0.472 & 0.511 & 0.333 & \textbf{0.562} & \underline{0.534} & 0.492 \\
& S$\uparrow$  & 0.468 & 0.516 & 0.428 & \textbf{0.540} & 0.473 & \underline{0.531} \\
& F1$\uparrow$ & 0.462 & 0.494 & 0.320 & \textbf{0.537} & 0.454 & \underline{0.495} \\
\midrule
\multirow{3}{*}{Pathology}
& P$\uparrow$  & 0.715 & \underline{0.819} & 0.790 & 0.808 & 0.797 & \textbf{0.838} \\
& S$\uparrow$  & 0.726 & \underline{0.797} & 0.767 & 0.781 & 0.791 & \textbf{0.837} \\
& F1$\uparrow$ & 0.717 & \underline{0.798} & 0.766 & 0.786 & 0.793 & \textbf{0.832} \\
\midrule
\multirow{3}{*}{Histology}
& P$\uparrow$  & 0.401 & \underline{0.472} & 0.419 & 0.456 & 0.421 & \textbf{0.502} \\
& S$\uparrow$  & 0.509 & \textbf{0.589} & 0.570 & \underline{0.576} & \underline{0.576} & 0.470 \\
& F1$\uparrow$ & 0.432 & \textbf{0.509} & 0.477 & \underline{0.498} & 0.484 & 0.451 \\
\bottomrule
\end{tabular*}
\end{table}
\subsubsection{Natural language generation.} We compare BUSTR with five representative report-generation methods: R2Gen~\cite{chen2020generating}, R2GenCMN~\cite{chen2021cross}, TSGET~\cite{yi2024tsget}, R2GenGPT~\cite{wang2023r2gengpt}, and the ultrasound report-generation method of Li et al.~\cite{li2024ultrasound}. The first four methods were originally developed primarily for chest radiograph report generation. Table~\ref{tab_nlg} summarizes the NLG performance. BUSTR achieves the highest reported NLG scores among the compared methods on both datasets. On BrEaST, it obtains BLEU-4, ROUGE-L, METEOR, and CIDEr scores of 0.418, 0.554, 0.337, and 0.625, respectively. On BUS-BRA, BUSTR achieves corresponding scores of 0.516, 0.658, 0.398, and 2.062. These results indicate that the combination of descriptor-aware visual supervision and the dual-level vision--language objective improves similarity to the descriptor-derived reports.

\subsubsection{Clinical efficacy.}
Tables~\ref{tab_ce_breast} and~\ref{tab_ce_busbra} report CE results for descriptor recovery from the generated reports. On BrEaST, BUSTR achieves the highest sensitivity and F1-score for BI-RADS. It also achieves the highest precision, sensitivity, and F1-score for lesion shape, margin, and posterior features. Echogenicity remains more challenging: BUSTR obtains a sensitivity of 0.560, while several baselines achieve higher precision or F1-score. On BUS-BRA, BUSTR achieves the highest precision, sensitivity, and F1-score for pathology. It also obtains the highest precision for histology; however, this improvement is accompanied by lower sensitivity, resulting in an F1-score below the strongest histology baselines. For BI-RADS, BUSTR achieves the second-highest sensitivity and F1-score but does not obtain the highest precision. Overall, the CE results show the clearest improvements for pathology and selected BrEaST lesion descriptors, while performance for BI-RADS, echogenicity, and histology remains more variable.

\begin{figure}[t]
\centering
\includegraphics[width=0.97\textwidth]{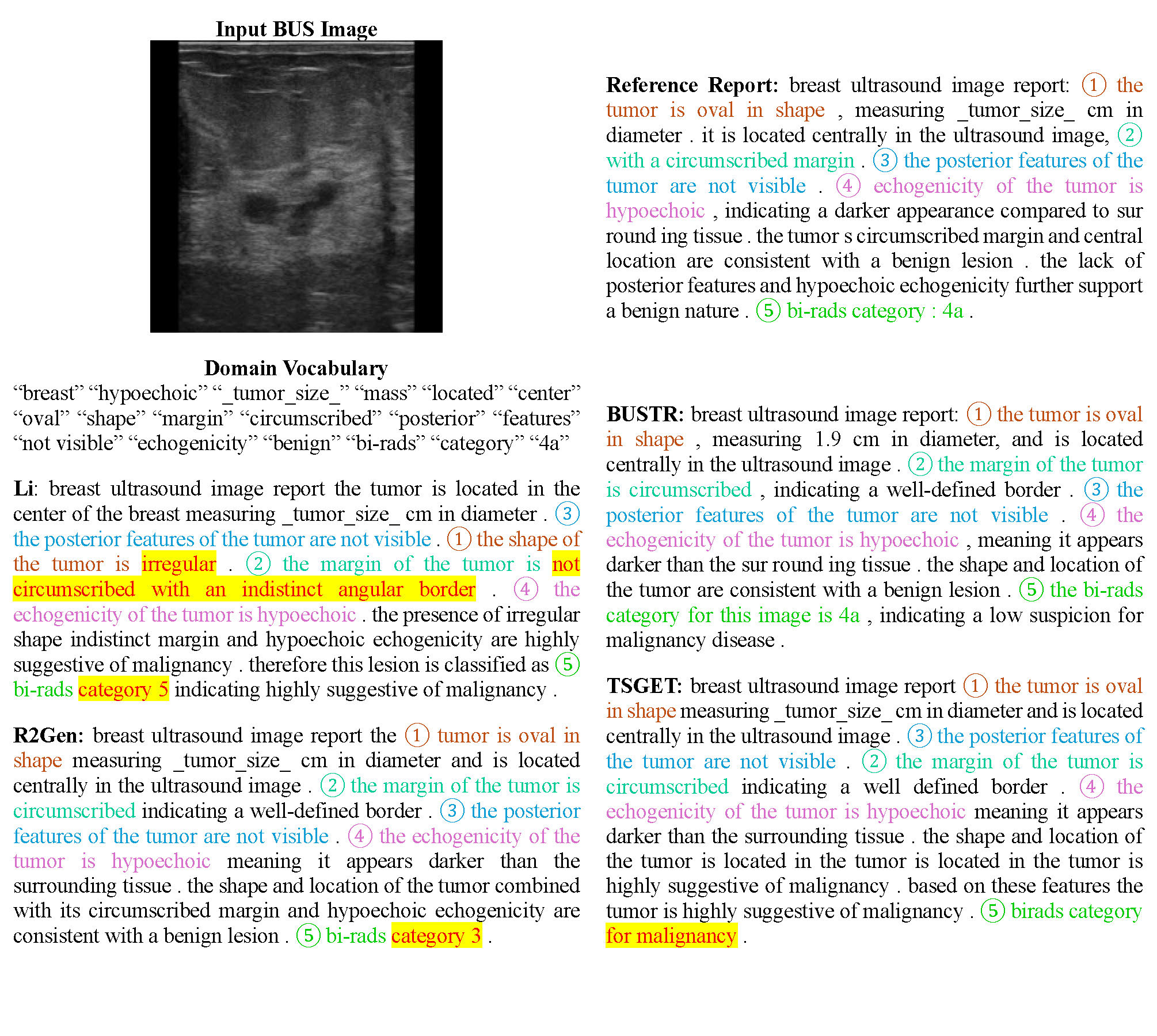}
\caption{Qualitative comparison of reports generated by the
top-performing methods. Colors identify the evaluated lesion descriptors,
and highlighted text indicates incorrect descriptor content relative to
the descriptor-derived reference report.}
\label{fig:reports_fig}
\end{figure}
\subsection{Statistical Analysis} To assess whether the observed performance gains are consistent across folds, we perform paired two-sided $t$-tests on the five-fold test scores for BLEU-4, ROUGE-L, METEOR, and CIDEr. For each dataset, BUSTR is compared with the baseline that achieved the highest CIDEr score in Table~\ref{tab_nlg}: R2Gen for BrEaST and R2GenGPT for BUS-BRA. On BrEaST, BUSTR significantly improves ROUGE-L, METEOR, and CIDEr over R2Gen ($p < 0.05$), while the improvement in BLEU-4 is positive but not statistically significant ($p \approx 0.14$). On BUS-BRA, BUSTR achieves statistically significant improvements over R2GenGPT for all four evaluated metrics ($p < 0.01$ in each case). These fold-level tests provide supportive evidence that the observed improvements over the selected comparison methods are consistent across the five folds. Given the limited number of folds and the evaluation of multiple metrics, the statistical results should be interpreted as supportive rather than definitive. \subsection{Ablation Study} To examine the contribution of each component, we conduct the ablation and sensitivity experiments summarized in Table~\ref{tab_nlg_ablation}. The following variants are evaluated: \begin{itemize} \item \textbf{Base Model:} a pre-trained Swin Transformer vision encoder with a single linear projection into the language-model embedding space and a standard token-level cross-entropy objective. \item \textbf{Vision Only:} the base model in which the vision encoder is replaced by the descriptor-aware multi-head encoder, while the training objective remains token-level cross-entropy. \item \textbf{Loss Only:} the base vision encoder trained using the dual-level objective that combines token-level cross-entropy with cosine-similarity alignment between the LLM hidden states and input embeddings. \item \textbf{Full BUSTR:} the complete model combining the descriptor-aware multi-head vision encoder with the dual-level objective. Sum, mean, maximum, and multiplication aggregation rules are evaluated for combining $\mathcal{L}_{\text{CE}}$ and $\mathcal{L}_{\text{Align}}$, together with multiple multitask loss-weight configurations. \end{itemize} Vision~1--4 denote the multitask loss-weight configurations used in the sensitivity analysis. These experiments assess how task weighting affects report-generation performance across datasets with partially overlapping annotation sets. The BrEaST and BUS-BRA configurations are reported in Tables~\ref{tab:weight_configs} and \ref{tab:busbra_weight_configs}, respectively. The results show that both the descriptor-aware vision encoder and the dual-level objective contribute to report-generation performance. On BrEaST, replacing the base encoder with the descriptor-aware multi-head encoder (Vision Only) improves BLEU-4 from 0.389 to 0.396, while applying only the dual-level loss (Loss Only) increases BLEU-4 to 0.400. On BUS-BRA, Vision Only improves BLEU-4 from 0.483 to 0.495, while Loss Only improves CIDEr from 1.924 to 2.087. When both components are combined, the BUSTR variants achieve strong overall performance. Among the mean-aggregation configurations, Vision~2--Mean provides the strongest combined BLEU-4, ROUGE-L, and METEOR performance on BrEaST, with scores of 0.418, 0.554, and 0.337, respectively. On BUS-BRA, Vision~4--Mean provides the strongest combined performance on these metrics, with scores of 0.516, 0.658, and 0.398. These configurations are therefore reported as the default BUSTR variants in the main comparison. Some alternative configurations achieve higher values on individual metrics, particularly CIDEr, as shown in Table~\ref{tab_nlg_ablation}. The sensitivity results indicate that the relative scaling and aggregation of $\mathcal{L}_{\text{CE}}$ and $\mathcal{L}_{\text{Align}}$ affect optimization and report-generation performance. The mean aggregation provides a balanced result across the principal NLG metrics and is therefore used as the default configuration.

\begin{table}[t]
\centering
\caption{Ablation and sensitivity analysis using NLG metrics. Bold indicates the best result, and underlining indicates the second-best result. Tied values are marked consistently.}
\label{tab_nlg_ablation}
\scriptsize
\renewcommand{\arraystretch}{1.08}
\setlength{\tabcolsep}{0.35em}
\begin{tabular*}{\textwidth}{@{\extracolsep{\fill}}llccccccc}
\toprule
Dataset & Approach & BLEU-1$\uparrow$ & BLEU-2$\uparrow$ & BLEU-3$\uparrow$ & BLEU-4$\uparrow$ & ROUGE-L$\uparrow$ & METEOR$\uparrow$ & CIDEr$\uparrow$ \\
\midrule
\multirow{19}{*}{BrEaST}
& Base Model      & 0.635 & 0.523 & 0.446 & 0.389 & 0.490 & 0.330 & 0.541 \\
& Vision Only     & 0.640 & 0.528 & 0.452 & 0.396 & 0.490 & 0.333 & 0.605 \\
& Loss Only       & 0.659 & 0.544 & 0.461 & 0.400 & 0.543 & 0.329 & 0.573 \\
\cmidrule(lr){2-9}
& Vision 1 - Sum  & \textbf{0.672} & \underline{0.556} & 0.474 & 0.413 & \textbf{0.555} & \textbf{0.338} & \underline{0.706} \\
& Vision 2 - Sum  & 0.667 & 0.551 & 0.471 & 0.411 & 0.551 & 0.336 & 0.689 \\
& Vision 3 - Sum  & 0.659 & 0.547 & 0.466 & 0.406 & 0.547 & 0.331 & 0.665 \\
& Vision 4 - Sum  & 0.653 & 0.537 & 0.452 & 0.390 & 0.542 & 0.325 & 0.499 \\
\cmidrule(lr){2-9}
& Vision 1 - Mean & 0.669 & 0.553 & 0.472 & 0.412 & 0.552 & 0.336 & 0.630 \\
& Vision 2 - Mean & \underline{0.671} & \textbf{0.558} & \textbf{0.478} & \textbf{0.418} & \underline{0.554} & \underline{0.337} & 0.625 \\
& Vision 3 - Mean & 0.669 & 0.552 & 0.469 & 0.407 & 0.546 & 0.333 & 0.655 \\
& Vision 4 - Mean & 0.660 & 0.543 & 0.460 & 0.399 & 0.551 & 0.333 & 0.599 \\
\cmidrule(lr){2-9}
& Vision 1 - Max  & 0.642 & 0.522 & 0.440 & 0.380 & 0.535 & 0.320 & 0.482 \\
& Vision 2 - Max  & 0.631 & 0.518 & 0.439 & 0.382 & 0.535 & 0.320 & 0.458 \\
& Vision 3 - Max  & 0.648 & 0.532 & 0.452 & 0.392 & 0.542 & 0.323 & 0.513 \\
& Vision 4 - Max  & 0.642 & 0.525 & 0.443 & 0.382 & 0.542 & 0.321 & 0.497 \\
\cmidrule(lr){2-9}
& Vision 1 - Mul  & 0.661 & 0.547 & 0.465 & 0.405 & 0.552 & 0.332 & 0.601 \\
& Vision 2 - Mul  & \textbf{0.672} & \textbf{0.558} & \underline{0.476} & \underline{0.416} & 0.553 & 0.336 & \textbf{0.719} \\
& Vision 3 - Mul  & 0.668 & 0.553 & 0.470 & 0.409 & \underline{0.554} & 0.335 & 0.653 \\
& Vision 4 - Mul  & 0.656 & 0.541 & 0.461 & 0.401 & 0.549 & 0.327 & 0.564 \\
\midrule
\multirow{19}{*}{BUS-BRA}
& Base Model      & 0.717 & 0.615 & 0.540 & 0.483 & 0.608 & 0.392 & 1.924 \\
& Vision Only     & 0.724 & 0.624 & 0.551 & 0.495 & 0.622 & \textbf{0.399} & 1.971 \\
& Loss Only       & 0.728 & 0.628 & 0.555 & 0.498 & 0.643 & 0.392 & 2.087 \\
\cmidrule(lr){2-9}
& Vision 1 - Sum  & \underline{0.736} & \underline{0.640} & 0.569 & 0.513 & 0.655 & \underline{0.398} & 2.072 \\
& Vision 2 - Sum  & 0.731 & 0.633 & 0.562 & 0.506 & 0.650 & 0.395 & 2.028 \\
& Vision 3 - Sum  & 0.733 & 0.637 & 0.567 & 0.511 & 0.652 & 0.394 & 2.014 \\
& Vision 4 - Sum  & 0.732 & 0.638 & 0.569 & 0.514 & \underline{0.656} & 0.395 & 1.981 \\
\cmidrule(lr){2-9}
& Vision 1 - Mean & \underline{0.736} & \underline{0.640} & \underline{0.570} & \underline{0.515} & 0.655 & \underline{0.398} & 2.099 \\
& Vision 2 - Mean & 0.733 & 0.637 & 0.567 & 0.511 & 0.653 & 0.395 & 2.043 \\
& Vision 3 - Mean & \underline{0.736} & 0.639 & 0.569 & 0.513 & 0.655 & 0.397 & \underline{2.102} \\
& Vision 4 - Mean & 0.735 & \textbf{0.641} & \textbf{0.571} & \textbf{0.516} & \textbf{0.658} & \underline{0.398} & 2.062 \\
\cmidrule(lr){2-9}
& Vision 1 - Max  & 0.724 & 0.632 & 0.564 & 0.511 & \underline{0.656} & 0.390 & 1.778 \\
& Vision 2 - Max  & 0.721 & 0.628 & 0.560 & 0.506 & 0.652 & 0.389 & 1.786 \\
& Vision 3 - Max  & 0.722 & 0.630 & 0.563 & 0.510 & 0.654 & 0.388 & 1.740 \\
& Vision 4 - Max  & 0.718 & 0.626 & 0.558 & 0.504 & 0.652 & 0.387 & 1.705 \\
\cmidrule(lr){2-9}
& Vision 1 - Mul  & \textbf{0.737} & \textbf{0.641} & \textbf{0.571} & \underline{0.515} & 0.655 & \underline{0.398} & 2.064 \\
& Vision 2 - Mul  & 0.732 & 0.638 & 0.568 & 0.513 & \underline{0.656} & 0.396 & 1.980 \\
& Vision 3 - Mul  & 0.735 & 0.638 & 0.567 & 0.511 & 0.653 & 0.396 & 2.075 \\
& Vision 4 - Mul  & 0.733 & 0.638 & 0.568 & 0.512 & 0.654 & 0.395 & \textbf{2.104} \\
\bottomrule
\end{tabular*}
\end{table}

\begin{table}[t]
\centering
\caption{Multitask loss-weight configurations used in the sensitivity analysis on the BrEaST dataset.}
\label{tab:weight_configs}
\small
\renewcommand{\arraystretch}{1.15}
\setlength{\tabcolsep}{0.75em}
\begin{tabular}{lcccccc}
\toprule
Model & Tumor Size & BI-RADS & Shape & Margin & Echo & Posterior \\
\midrule
Vision 1 & 16.7\% & 16.7\% & 16.7\% & 16.7\% & 16.7\% & 16.7\% \\
Vision 2 & 20\% & 20\% & 15\% & 15\% & 15\% & 15\% \\
Vision 3 & 15\% & 25\% & 15\% & 15\% & 15\% & 15\% \\
Vision 4 & 10\% & 50\% & 10\% & 10\% & 10\% & 10\% \\
\bottomrule
\end{tabular}
\end{table}

\begin{table}[t]
\centering
\caption{Multitask loss-weight configurations used in the sensitivity analysis on the BUS-BRA dataset.}
\label{tab:busbra_weight_configs}
\small
\renewcommand{\arraystretch}{1.15}
\setlength{\tabcolsep}{0.65em}
\begin{tabular}{lcccc}
\toprule
Model & Tumor Size & BI-RADS & Histology & Pathology \\
\midrule
Vision 1 & 25\% & 25\% & 25\% & 25\% \\
Vision 2 & 20\% & 40\% & 20\% & 20\% \\
Vision 3 & 30\% & 30\% & 20\% & 20\% \\
Vision 4 & 40\% & 40\% & 10\% & 10\% \\
\bottomrule
\end{tabular}
\end{table}

\section{Discussion}

This study demonstrates that structured BUS descriptors, lesion masks, and radiomics features can provide useful supervision for descriptor-aware breast ultrasound report generation. BUSTR achieved the highest NLG scores among the compared methods on both datasets (Table~\ref{tab_nlg}). The paired two-sided $t$-tests further support these gains against the selected comparison methods. On BrEaST, BUSTR significantly improved ROUGE-L, METEOR, and CIDEr over R2Gen ($p < 0.05$), while the improvement in BLEU-4 was positive but not statistically significant ($p \approx 0.14$). On BUS-BRA, BUSTR significantly improved all four evaluated NLG metrics over R2GenGPT ($p < 0.01$). These results suggest that descriptor-derived reports can provide a consistent report-level training signal when paired radiologist-written reports are unavailable.

The observed performance gains appear to arise from both the descriptor-aware vision encoder and the dual-level vision--language training objective. The multi-head Swin encoder learns BUS image representations through explicit supervision from lesion descriptors and auxiliary targets, thereby aligning visual features with clinically meaningful structured information. The ablation study (Table~\ref{tab_nlg_ablation}) shows that replacing the base encoder with the descriptor-aware multi-head encoder improves NLG performance, supporting the value of descriptor-level visual supervision. Similarly, the Loss Only variant improves performance over the base model, indicating that the cosine-similarity alignment term helps maintain consistency between image-conditioned inputs and generated report representations. When both components are combined, the full BUSTR variants achieve the strongest overall performance. The sensitivity analysis further shows that report-generation performance depends on both the multitask loss weights and the method used to combine $\mathcal{L}_{\text{CE}}$ and $\mathcal{L}_{\text{Align}}$. The selected mean-based configurations provide a balanced performance across the main NLG metrics.

The CE results provide a complementary assessment of descriptor-level factual consistency. On BrEaST, BUSTR achieved the highest sensitivity and F1-score for BI-RADS and the strongest overall performance for lesion shape, margin, and posterior features (Table~\ref{tab_ce_breast}). Echogenicity remained more challenging, with several baselines achieving higher precision or F1-score. On BUS-BRA, BUSTR achieved the best precision, sensitivity, and F1-score for pathology and the highest precision for histology (Table~\ref{tab_ce_busbra}). However, the increased histology precision was accompanied by lower sensitivity, resulting in an F1-score below the strongest baselines. BUSTR also achieved competitive BI-RADS sensitivity on BUS-BRA but did not obtain the highest BI-RADS precision or F1-score. These findings suggest that the method is most effective for descriptors that are sufficiently represented and visually distinguishable, while performance remains more variable for imbalanced, subtle, or heterogeneous categories.

The qualitative examples in Fig.~\ref{fig:reports_fig} further illustrate that BUSTR produces more coherent and structured lesion descriptions than the compared baselines, with fewer errors in several key descriptor fields. However, these examples should be interpreted as evidence of improved descriptor-aware report generation rather than evidence of complete clinical report quality. Tumor-size estimation remains challenging because it is a continuous quantity affected by lesion-mask quality, image scaling, and boundary uncertainty. BUSTR therefore uses a dedicated regression head to estimate tumor size and inserts the resulting value into the generated report rather than relying on the language model alone to infer it. Nevertheless, the clinical accuracy of the predicted size requires further quantitative and expert evaluation.

Because LLM-based report generation can introduce unsupported findings, the CE metrics are also used as a partial measure of descriptor-level factual consistency. The main observed errors involved incorrect or missing recovery of BI-RADS category, lesion shape, margin, echogenicity, posterior features, pathology, histology, or tumor size. The constrained construction of descriptor-derived reports is intended to reduce unsupported content in the supervisory text, while the CE evaluation measures whether selected structured facts are preserved in the generated reports. However, these mechanisms do not eliminate hallucinations and do not capture unsupported statements outside the predefined descriptor set. Future work should therefore include a dedicated hallucination analysis that identifies both omitted descriptors and findings not supported by the source image or structured annotations, together with radiologist review.

An important implication of this work is that structured annotations can support more than classification or segmentation. They can also provide intermediate supervision for learning clinically meaningful visual representations and generating organized report text. This is particularly relevant for BUS, where public datasets often include BI-RADS descriptors, pathology labels, histology labels, lesion masks, or related metadata, but not paired radiologist-written reports. BUSTR addresses this constrained but practical setting by connecting BUS image features, structured lesion descriptors, radiomics information, and report generation within a unified descriptor-aware framework.

This study has several limitations. First, the evaluation is based on two public BUS datasets, BrEaST and BUS-BRA, which have limited sample sizes, incomplete annotations, and imbalanced descriptor distributions. BrEaST is particularly small, and neither dataset provides paired radiologist-written reports. Consequently, the descriptor-derived reports do not capture the full linguistic diversity, diagnostic reasoning, uncertainty expression, or contextual detail of expert-authored clinical reports. Their language may also reflect stylistic patterns introduced by the LLM used during reference-report construction. Second, the generated reports are restricted by the available descriptors and radiomics features; findings, clinical context, and terminology absent from the datasets cannot be reliably generated or evaluated. Third, the study does not include an independent external dataset, and the generalizability of the framework across institutions, ultrasound systems, and patient populations remains uncertain. Fourth, the evaluation relies primarily on automatic NLG and CE metrics. These metrics measure report similarity and descriptor recovery but do not fully assess clinical usefulness, readability, trust, factual completeness, or workflow impact. Tumor-size accuracy and unsupported-finding rates were also not evaluated as dedicated outcomes. Finally, the study does not include a radiologist-in-the-loop reader assessment. Future work should evaluate BUSTR on larger and more diverse BUS datasets, incorporate paired expert-authored reports when available, include dedicated factuality and size-estimation analyses, and assess the clinical usefulness of generated reports through expert review.

\section{Conclusion}

We presented BUSTR, a descriptor-aware vision--language framework for breast ultrasound report generation. BUSTR constructs descriptor-derived reports from structured lesion annotations and radiomics features, learns descriptor-aware visual representations using a multi-head Swin Transformer with multitask supervision, and connects visual and textual representations through a dual-level objective combining token-level cross-entropy with representation-level cosine alignment. Experiments on the BrEaST and BUS-BRA datasets show that BUSTR improves NLG performance compared with representative report-generation baselines. It also improves descriptor recovery for clinically relevant targets, particularly lesion shape, margin, posterior features, and pathology, while achieving improved BI-RADS sensitivity and F1-score on BrEaST.

These findings suggest that structured BUS descriptors, lesion masks, and radiomics features can provide useful supervision for report generation when paired radiologist-written reports are unavailable. BUSTR demonstrates a practical strategy for using structured annotations beyond classification and segmentation by converting them into report-level supervision for image-conditioned text generation. The generated reports may support structured reporting, educational drafting, dataset documentation, and radiologist review, but they are not intended to replace expert-authored clinical reports or autonomous diagnostic decision-making. Future work will evaluate the framework on larger and more diverse datasets, incorporate paired radiology reports when available, and include radiologist-in-the-loop studies to assess factuality, clinical usefulness, readability, trust, and workflow integration.

\section*{Acknowledgments}
Acknowledgments will be added after peer review.

\section*{Ethics statement}
This study used publicly available, de-identified breast ultrasound datasets and did not involve direct interaction with human participants or access to identifiable private information. Therefore, formal ethics approval and informed consent were not required for this study.

\section*{Data availability}
The datasets analyzed in this study are publicly available from the original BrEaST and BUS-BRA dataset sources cited in the manuscript. The source code will be made available upon publication.

\section*{Declaration of competing interest}
The authors declare that they have no known competing financial interests or personal relationships that could have appeared to influence the work reported in this paper.

\section*{Declaration of generative AI and AI-assisted technologies in the writing process}
During the preparation of this work, the authors used ChatGPT to assist with language editing, organization, and readability. After using this tool, the authors reviewed and edited the content as needed and take full responsibility for the content of the submitted manuscript.

\bibliographystyle{unsrtnat}
\bibliography{references}

@inproceedings{ramesh2022improving,
  title={Improving radiology report generation systems by removing hallucinated references to non-existent priors},
  author={Ramesh, Vignav and Chi, Nathan A and Rajpurkar, Pranav},
  booktitle={Machine Learning for Health},
  pages={456--473},
  year={2022},
  organization={PMLR}
}

@article{sun2025challenge,
  title={Challenge-aware U-net for breast lesion segmentation in ultrasound images},
  author={Sun, Dengdi and Dong, Changxu and Yan, Yuchen and Jiang, Bo and Duan, Yayang and Tu, Zhengzheng and Zhang, Chaoxue},
  journal={Pattern Recognition},
  pages={111851},
  year={2025},
  publisher={Elsevier}
}

@article{mahesh2025enhancing,
  title={Enhancing diagnostic precision in breast cancer classification through EfficientNetB7 using advanced image augmentation and interpretation techniques},
  author={Mahesh, TR and Khan, Surbhi Bhatia and Mishra, Kritika Kumari and Alzahrani, Saeed and Alojail, Mohammed},
  journal={International Journal of Imaging Systems and Technology},
  volume={35},
  number={1},
  pages={e70000},
  year={2025},
  publisher={Wiley Online Library}
}

@ARTICLE{ChenAlign,
  author={Chen, Haoyuan and Li, Yonghao and Zhang, Jiadong and Yang, Long and Sun, Yiqun and Chen, Yaling and Zhou, Shichong and Li, Zhenhui and Qian, Xuejun and Xu, Qi and Shen, Dinggang},
  journal={IEEE Transactions on Medical Imaging}, 
  title={An Alignment and Imputation Network (AINet) for Breast Cancer Diagnosis with Multimodal Multi-view Ultrasound Images}, 
  year={2025},
  volume={},
  number={},
  pages={1-1},
  doi={10.1109/TMI.2025.3625254}}

@INPROCEEDINGS{BCS-NET,
  author={Li, Ruili and Li, Ruiyu and Takaya, Eichi and Lin, Zizhen and Kobayashi, Tomoya and Mtsuda, Nanako and Ueda, Takuya},
  booktitle={ICASSP 2025 - 2025 IEEE International Conference on Acoustics, Speech and Signal Processing (ICASSP)}, 
  title={BCS-Net: Multi-Task Breast Cancer Screening Network Enhanced by Multi-Modality Attention}, 
  year={2025},
  volume={},
  number={},
  pages={1-5},
  doi={10.1109/ICASSP49660.2025.10888652}}

@article{ge2023ai,
  title={AI-assisted Method for Efficiently Generating Breast Ultrasound Screening Reports},
  author={Ge, Shuang and Ye, Qiongyu and Xie, Wenquan and Sun, Desheng and Zhang, Huabin and Zhou, Xiaobo and Yuan, Kehong},
  journal={Current Medical Imaging},
  volume={19},
  number={2},
  pages={149--157},
  year={2023}
}

@inproceedings{tommasi2008svm,
  title={An SVM confidence-based approach to medical image annotation},
  author={Tommasi, Tatiana and Orabona, Francesco and Caputo, Barbara},
  booktitle={Workshop of the Cross-Language Evaluation Forum for European Languages},
  pages={696--703},
  year={2008},
  organization={Springer}
}

@inproceedings{gobeill2009query,
  title={Query and document expansion with medical subject headings terms at medical imageclef 2008},
  author={Gobeill, Julien and Ruch, Patrick and Zhou, Xin},
  booktitle={Evaluating Systems for Multilingual and Multimodal Information Access: 9th Workshop of the Cross-Language Evaluation Forum, CLEF 2008, Aarhus, Denmark, September 17-19, 2008, Revised Selected Papers 9},
  pages={736--743},
  year={2009},
  organization={Springer}
}

@article{shareef2022estan,
  title   = {ESTAN: Enhanced Small Tumor-Aware Network for Breast Ultrasound Image Segmentation},
  author  = {Shareef, Bryar and Vakanski, Aleksandar and Freer, Phoebe E. and Xian, Min},
  journal = {Healthcare},
  volume  = {10},
  number  = {11},
  pages   = {2262},
  year    = {2022},
  doi     = {10.3390/healthcare10112262}
}

@article{azhar2024_bus_reports,
  author  = {Azhar, Khadija and Lee, Byoung-Dai and Byon, Shi Sub
             and Cho, Kyu Ran and Song, Sung Eun},
  title   = {AI-Powered Synthesis of Structured Multimodal Breast Ultrasound
             Reports Integrating Radiologist Annotations and Deep Learning Analysis},
  journal = {Bioengineering},
  year    = {2024},
  volume  = {11},
  number  = {9},
  pages   = {890},
  doi     = {10.3390/bioengineering11090890}
}

@article{huh2025wholistic,
  title={Wholistic report generation for Breast ultrasound using LangChain},
  author={Huh, Jaeyoung and Ahn, Hye Shin and Park, Hyun Jeong and Ye, Jong Chul},
  journal={Computerized Medical Imaging and Graphics},
  pages={102697},
  year={2025},
  publisher={Elsevier}
}

@article{wang2025survey,
  title={A survey of deep-learning-based radiology report generation using multimodal inputs},
  author={Wang, Xinyi and Figueredo, Grazziela and Li, Ruizhe and Zhang, Wei Emma and Chen, Weitong and Chen, Xin},
  journal={Medical Image Analysis},
  pages={103627},
  year={2025},
  publisher={Elsevier}
}

@inproceedings{wang2023_metransformer,
  author    = {Wang, Zhanyu and Liu, Lingqiao and Wang, Lei and Zhou, Luping},
  title     = {{METransformer}: Radiology Report Generation by Transformer
               with Multiple Learnable Expert Tokens},
  booktitle = {Proceedings of the IEEE/CVF Conference on Computer Vision and Pattern Recognition (CVPR)},
  year      = {2023},
  pages     = {16302--16312}
}

@inproceedings{gu2025radalign,
  title={Radalign: Advancing radiology report generation with vision-language concept alignment},
  author={Gu, Difei and Gao, Yunhe and Zhou, Yang and Zhou, Mu and Metaxas, Dimitris},
  booktitle={International Conference on Medical Image Computing and Computer-Assisted Intervention},
  pages={484--494},
  year={2025},
  organization={Springer}
}

@article{parres2024_rrg_rl,
  author  = {Parres, Daniel and Albiol, Alberto and Paredes, Roberto},
  title   = {Improving Radiology Report Generation Quality and Diversity through
             Reinforcement Learning and Text Augmentation},
  journal = {Bioengineering},
  year    = {2024},
  volume  = {11},
  number  = {4},
  pages   = {351},
  doi     = {10.3390/bioengineering11040351}
}

@article{margolies2016_bigdata,
  author  = {Margolies, Laurie R. and Pandey, Gaurav
             and Horowitz, Eliot R. and Mendelson, David S.},
  title   = {Breast Imaging in the Era of Big Data: Structured Reporting and Data Mining},
  journal = {AJR American Journal of Roentgenology},
  year    = {2016},
  volume  = {206},
  number  = {2},
  pages   = {259--264},
  doi     = {10.2214/AJR.15.15396}
}

@inproceedings{shareef2023hybrid,
  title     = {Breast Ultrasound Tumor Classification Using a Hybrid Multitask CNN--Transformer Network},
  author    = {Shareef, Bryar and Xian, Min and Vakanski, Aleksandar and Wang, Haotian},
  booktitle = {Medical Image Computing and Computer-Assisted Intervention (MICCAI)},
  year      = {2023},
  doi       = {10.1007/978-3-031-43901-8_33}
}

@article{busis,
  author = {Zhang, Y. and Xian, M. and Cheng, H.D. and Shareef, B. and Ding, J. and Xu, F. and Huang, K. and Zhang, B. and Ning, C. and Wang, Y.},
  title = {BUSIS: A Benchmark for Breast Ultrasound Image Segmentation},
  journal = {Healthcare (Basel)},
  volume = {10},
  number = {4},
  pages = {729},
  year = {2022},
  doi = {10.3390/healthcare10040729},
  pmid = {35455906},
  pmcid = {PMC9025635}
}

@article{li2024ultrasound,
  title={Ultrasound Report Generation with Cross-Modality Feature Alignment via Unsupervised Guidance},
  author={Li, Jun and Su, Tongkun and Zhao, Baoliang and Lv, Faqin and Wang, Qiong and Navab, Nassir and Hu, Ying and Jiang, Zhongliang},
  journal={IEEE Transactions on Medical Imaging},
  year={2024},
  publisher={IEEE}
}

@article{wang2023r2gengpt,
  author = "{Wang, Z. and Liu, L. and Wang, L. and Zhou, L.}",
  title = {R2GenGPT: Radiology report generation with frozen LLMs},
  journal = {Meta-Radiology},
  volume = {1},
  number = {3},
  pages = {100033},
  year = {2023},
  doi = {10.1016/j.metrad.2023.100033}
}

@inproceedings{pal2024gemini,
  title={Gemini goes to med school: exploring the capabilities of multimodal large language models on medical challenge problems \& hallucinations},
  author={Pal, Ankit and Sankarasubbu, Malaikannan},
  booktitle={Proceedings of the 6th Clinical Natural Language Processing Workshop},
  pages={21--46},
  year={2024}
}

@inproceedings{vedantam2015cider,
  title={Cider: Consensus-based image description evaluation},
  author={Vedantam, Ramakrishna and Lawrence Zitnick, C and Parikh, Devi},
  booktitle={Proceedings of the IEEE conference on computer vision and pattern recognition},
  pages={4566--4575},
  year={2015}
}

@inproceedings{papineni2002bleu,
  title={Bleu: a method for automatic evaluation of machine translation},
  author={Papineni, Kishore and Roukos, Salim and Ward, Todd and Zhu, Wei-Jing},
  booktitle={Proceedings of the 40th annual meeting of the Association for Computational Linguistics},
  pages={311--318},
  year={2002}
}

@inproceedings{lin2004rouge,
  title={Rouge: A package for automatic evaluation of summaries},
  author={Lin, Chin-Yew},
  booktitle={Text summarization branches out},
  pages={74--81},
  year={2004}
}

@inproceedings{banerjee2005meteor,
  title={METEOR: An automatic metric for MT evaluation with improved correlation with human judgments},
  author={Banerjee, Satanjeev and Lavie, Alon},
  booktitle={Proceedings of the acl workshop on intrinsic and extrinsic evaluation measures for machine translation and/or summarization},
  pages={65--72},
  year={2005}
}

@article{touvron2023llama,
  title={Llama 2: Open foundation and fine-tuned chat models},
  author={Touvron, Hugo and Martin, Louis and Stone, Kevin and Albert, Peter and Almahairi, Amjad and Babaei, Yasmine and Bashlykov, Nikolay and Batra, Soumya and Bhargava, Prajjwal and Bhosale, Shruti and others},
  journal={arXiv preprint arXiv:2307.09288},
  year={2023}
}

@article{yi2024tsget,
  title={TSGET: Two-stage global enhanced transformer for automatic radiology report generation},
  author={Yi, Xiulong and Fu, You and Liu, Ruiqing and Zhang, Hao and Hua, Rong},
  journal={IEEE Journal of Biomedical and Health Informatics},
  volume={28},
  number={4},
  pages={2152--2162},
  year={2024},
  publisher={IEEE}
}

@inproceedings{chen2020generating,
  title={Generating Radiology Reports via Memory-driven Transformer},
  author={Chen, Zhihong and Song, Yan and Chang, Tsung-Hui and Wan, Xiang},
  booktitle={Proceedings of the 2020 Conference on Empirical Methods in Natural Language Processing (EMNLP)},
  pages={1439--1449},
  year={2020}
}

@article{kolb2002comparison,
  title={Comparison of the performance of screening mammography, physical examination, and breast US and evaluation of factors that influence them: an analysis of 27,825 patient evaluations},
  author={Kolb, Thomas M and Lichy, Jacob and Newhouse, Jeffrey H},
  journal={Radiology},
  volume={225},
  number={1},
  pages={165--175},
  year={2002},
  publisher={Radiological Society of North America}
}

@techreport{markit2017complexities,
  author      = {{IHS Markit}},
  title       = {The Complexities of Physician Supply and Demand:
                 Projections from 2015 to 2030},
  institution = {Association of American Medical Colleges},
  year        = {2017}
}

@article{sun2023evaluating,
  title={Evaluating GPT-4 on impressions generation in radiology reports},
  author={Sun, Zhaoyi and Ong, Hanley and Kennedy, Patrick and Tang, Liyan and Chen, Shirley and Elias, Jonathan and Lucas, Eugene and Shih, George and Peng, Yifan},
  journal={Radiology},
  volume={307},
  number={5},
  pages={e231259},
  year={2023},
  publisher={Radiological Society of North America}
}

@article{lo2024automated,
  title={Automated breast imaging report generation based on the integration of multiple image features in a metadata format for shared decision-making},
  author={Lo, Chung-Ming and Chen, Hui-Ru},
  journal={Health informatics journal},
  volume={30},
  number={3},
  pages={14604582241288460},
  year={2024}
}

@article{qin2023computer,
  title={Computer-Aided Diagnosis System for Breast Ultrasound Reports Generation and Classification Method Based on Deep Learning},
  author={Qin, Haojun and Zhang, Lei and Guo, Quan},
  journal={Applied Sciences},
  volume={13},
  number={11},
  pages={6577},
  year={2023},
  publisher={MDPI}
}

@inproceedings{chen2021cross,
  title={Cross-modal Memory Networks for Radiology Report Generation},
  author={Chen, Zhihong and Shen, Yaling and Song, Yan and Wan, Xiang},
  booktitle={Proceedings of the 59th Annual Meeting of the Association for Computational Linguistics and the 11th International Joint Conference on Natural Language Processing (Volume 1: Long Papers)},
  pages={5904--5914},
  year={2021}
}

@article{zhang2023bi,
  title={BI-RADS-NET-V2: a composite multi-task neural network for computer-aided diagnosis of breast cancer in ultrasound images with semantic and quantitative explanations},
  author={Zhang, Boyu and Vakanski, Aleksandar and Xian, Min},
  journal={IEEE Access},
  volume={11},
  pages={79480--79494},
  year={2023},
  publisher={IEEE}
}

@article{sirshar2022attention,
  title={Attention based automated radiology report generation using CNN and LSTM},
  author={Sirshar, Mehreen and Paracha, Muhammad Faheem Khalil and Akram, Muhammad Usman and Alghamdi, Norah Saleh and Zaidi, Syeda Zainab Yousuf and Fatima, Tatheer},
  journal={Plos one},
  volume={17},
  number={1},
  pages={e0262209},
  year={2022},
  publisher={Public Library of Science San Francisco, CA USA}
}

@inproceedings{liu2021swin,
  title={Swin transformer: Hierarchical vision transformer using shifted windows},
  author={Liu, Ze and Lin, Yutong and Cao, Yue and Hu, Han and Wei, Yixuan and Zhang, Zheng and Lin, Stephen and Guo, Baining},
  booktitle={Proceedings of the IEEE/CVF international conference on computer vision},
  pages={10012--10022},
  year={2021}
}

@article{gomez2024bus,
  title={BUS-BRA: a breast ultrasound dataset for assessing computer-aided diagnosis systems},
  author={G{\'o}mez-Flores, Wilfrido and Gregorio-Calas, Maria Julia and Coelho de Albuquerque Pereira, Wagner},
  journal={Medical Physics},
  volume={51},
  number={4},
  pages={3110--3123},
  year={2024},
  publisher={Wiley Online Library}
}

@article{pawlowska2024curated,
  title={Curated benchmark dataset for ultrasound based breast lesion analysis},
  author={Paw{\l}owska, Anna and {\'C}wierz-Pie{\'n}kowska, Anna and Domalik, Agnieszka and Jagu{\'s}, Dominika and Kasprzak, Piotr and Matkowski, Rafa{\l} and Fura, {\L}ukasz and Nowicki, Andrzej and {\.Z}o{\l}ek, Norbert},
  journal={Scientific Data},
  volume={11},
  number={1},
  pages={148},
  year={2024},
  publisher={Nature Publishing Group UK London}
}

@inproceedings{park2025dart,
  title={DART: Disease-aware Image-Text Alignment and Self-correcting Re-alignment for Trustworthy Radiology Report Generation},
  author={Park, Sang-Jun and Heo, Keun-Soo and Shin, Dong-Hee and Son, Young-Han and Oh, Ji-Hye and Kam, Tae-Eui},
  booktitle={Proceedings of the Computer Vision and Pattern Recognition Conference},
  pages={15580--15589},
  year={2025}
}

@article{gao2025abnormal,
  title={Abnormal-region-aware Multi-modal Feature Fusion for medical report generation},
  author={Gao, Yan and Ni, Zhiwei and Liu, Wentao and Ni, Liping and Xin, Ling and Hu, Linbo and Zhang, Li},
  journal={Knowledge-Based Systems},
  pages={113538},
  year={2025},
  publisher={Elsevier}
}

@article{li2024organ,
  title={An organ-aware diagnosis framework for radiology report generation},
  author={Li, Shiyu and Qiao, Pengchong and Wang, Lin and Ning, Munan and Yuan, Li and Zheng, Yefeng and Chen, Jie},
  journal={IEEE Transactions on Medical Imaging},
  volume={43},
  number={12},
  pages={4253--4265},
  year={2024},
  publisher={IEEE}
}

@inproceedings{chen2024anatomy,
  title={Anatomy-Aware Enhancement and Cross-Modal Disease Representation Fusion for Medical Report Generation},
  author={Chen, Jianbin and Yang, Kai and Lin, Runfeng and Wang, Yating and Xu, Dacheng and Zhang, Meng and Liu, Shouqiang},
  booktitle={2024 IEEE International Conference on Systems, Man, and Cybernetics (SMC)},
  pages={315--320},
  year={2024},
  organization={IEEE}
}



\end{document}